\documentclass[lettersize,journal]{IEEEtran}
\usepackage{amsmath,amsfonts}
\usepackage{algorithmic}
\usepackage{algorithm}
\usepackage{array}
\usepackage[caption=false,font=normalsize,labelfont=sf,textfont=sf]{subfig}
\usepackage{textcomp}
\usepackage{stfloats}
\usepackage{url}
\usepackage{verbatim}
\usepackage{graphicx}
\usepackage{cite}
\usepackage{booktabs}
\usepackage{multirow}
\usepackage{color}
\usepackage{xcolor}
\usepackage{colortbl}
\begin{document}

\title{Seeing through Denoising: Frequency-guided One-step Denoising for Infrared Small Target Detection}

\author{
Xuelin Qian, Zixuan Wang, Chang Xu, Dingwen Zhang and Junwei Han}

\maketitle

\begin{abstract}
Infrared small target detection (IRSTD) is a challenging task due to the extremely small size of targets, low signal-to-noise ratios, and heavy background clutter in infrared imagery. While the progress of deep learning has advanced this field, existing methods primarily adopt discriminative paradigms that directly map images to target masks, overlooking the potential of generative modeling.
In this paper, we propose DiffIRSTD, a novel generative framework that reformulates infrared small target detection as a background reconstruction problem. Instead of explicitly predicting target regions, 
we reinterpret small targets as noise-like signals and leverage a diffusion-based model to recover a clean, target-free background from a perturbed input image. The residual between the reconstructed background and the original image is then exploited to highlight candidate target regions, providing explicit cues for subsequent detection.
To address the computational cost typically associated with diffusion sampling, we introduce a one-step denoising strategy that enables efficient background reconstruction while maintaining practical inference speed. Furthermore, we present a frequency-guided perception loss to preserve high-frequency background details and focus more on the reconstruction quality of target regions.
More importantly, our DiffIRSTD operates as a plug-and-play framework, compatible with existing IRSTD models without increasing the computational complexity of detectors.
Extensive experiments on IRSTD-1K, NUAA-SIRST, and NUDT-SIRST datasets demonstrate the superiority of our method, achieving state-of-the-art performance across three benchmarks.
\end{abstract}

\begin{IEEEkeywords}
Infrared small target detection, one-step denoising, background reconstruction, frequency-guided perceptual loss.
\end{IEEEkeywords}

\section{Introduction}

\IEEEPARstart{I}{nfrared} small target detection (IRSTD) aims to automatically localize and segment small targets in infrared imagery, where targets often occupy only a few pixels and lack clear structural or texture cues \cite{zhao2022single}. 
Such targets typically appear as dim and point-like targets, such as distant aircraft, missiles, or drones, in complex and cluttered backgrounds. IRSTD is critical in numerous military and civilian scenarios, including early warning systems, long-range surveillance, and autonomous sensing. However, the long imaging distance, low signal-to-noise ratio (SNR), and strong background interference often cause targets to be submerged in noise and clutter, making robust and accurate detection highly challenging.

Recent advances in deep learning have significantly improved IRSTD performance. Early approaches mainly rely on U-Net architectures, which enhance small target representation through multi-scale feature interaction \cite{ren2021dnanet,wu2022uiu,liu2024infrared,zhang2022isnet}. Transformer-based models are later introduced to capture long-range dependencies and global contextual information, further boosting detection accuracy \cite{yuan2024sctransnet,hu2025datransnet,zhang2022rkformer}. More recently, large vision foundation models, such as Mamba \cite{chen2024mim,zhang2025irmamba,jiang2025eamnet} and SAM \cite{zhang2024irsam,fu2025unified,zhang2025saist}, have been explored to enrich feature representations and improve generalization. Despite these developments, most existing methods remain fundamentally discriminative, focusing on learning direct mappings from images to target masks, while the potential of generative modeling in IRSTD remains underexplored.

\IEEEpubidadjcol
Diffusion models \cite{croitoru2023diffusion} have recently demonstrated remarkable capability in generative modeling and have achieved impressive success in image synthesis, restoration, and inpainting. Existing diffusion-based studies primarily explore two directions of either using diffusion models to generate additional training samples to alleviate data scarcity \cite{ding2024infrared}, or employing diffusion-based networks as iterative decoders to progressively refine predicted masks \cite{li2024dense,rong2026novel,chenhao2025diffusion}. 
Nevertheless, data generation struggles to faithfully synthesize small infrared targets due to their low intensity and ambiguous boundaries, whereas iterative diffusion decoders may introduce additional computational overhead, which limits deployment in resource-constrained scenarios.

\begin{figure}[t]
    \centering  
    \includegraphics[width=1\linewidth]{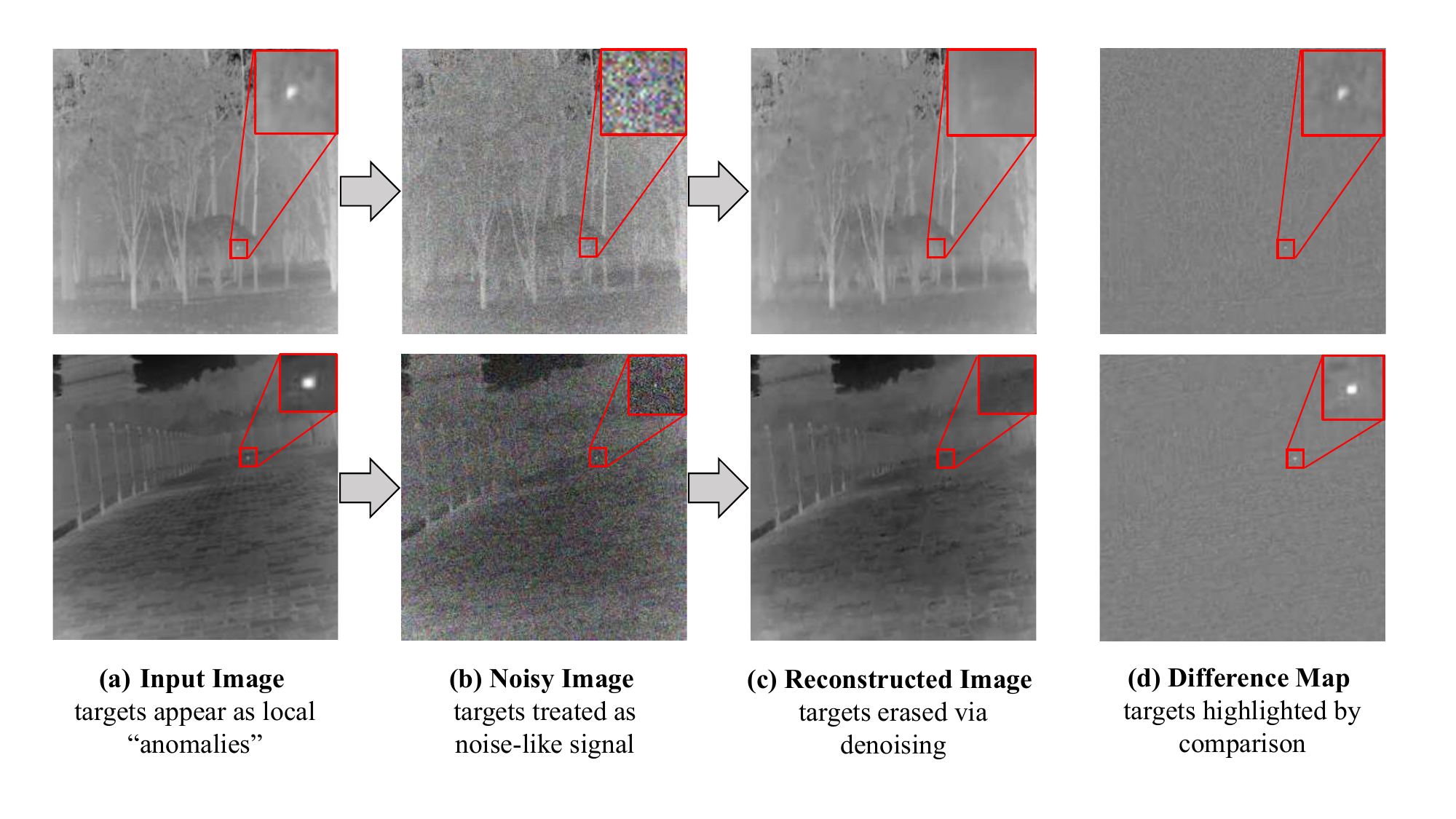}
    \vspace{-0.3in}
    \caption{\textbf{Illustration of our motivation.} From a new perspective of denoising, (a) we treat small targets as noise-like anomalies, and (b-c) then denoise the perturbed image to obtain a target-free background image. (d) The difference between the reconstructed background and the original image can be viewed as informative clues to highlight candidate target regions, promoting the subsequent detection accuracy.
    }
    \label{fig:teaser}
    \vspace{-0.2in}
\end{figure}

From an imaging perspective, infrared small targets typically exhibit intensity values that differ from their surrounding background, appearing as local ``anomalies'' against the smooth background (see Fig.~\ref{fig:teaser}(a)). However, factors such as blurry target boundaries, complex background textures, and low SNR frequently obscure these ``anomalies''. Motivated by this analysis, we propose a novel generative paradigm for infrared small target detection, termed \textbf{DiffIRSTD}. Instead of directly predicting targets from raw images, we reinterpret small targets as noise-like signals and leverage diffusion models to denoise them from the image, as illustrated in Fig.~\ref{fig:teaser}(b)-(c). By reconstructing a clean background image and comparing it with the original input, the target regions can be explicitly highlighted (Fig.~\ref{fig:teaser}(d)). This counterintuitive perspective transforms the detection problem into a background reconstruction problem. Reconstructing the low-rank background is expected to be easier than directly generating small target masks, and this pipeline is not required to alter the architecture or increase the complexity of the downstream detector.

Specifically, our DiffIRSTD framework consists of a reconstruction network for generating target-free background images and a detection network for localizing small targets. Given an input infrared image, we first perturb it by adding Gaussian noise, where the target regions are implicitly treated as part of the noise. The noisy image is then processed by a diffusion model to reconstruct a clean background image in which the targets are removed. 
Diffusion models typically require dozens or even hundreds of sampling steps to generate high-quality images. Such heavy iterative sampling is computationally expensive and conflicts with the efficiency requirements of infrared small target detection. To address this issue, we propose a one-step denoising strategy, which performs only a single denoising step to directly reconstruct the target-free background from the noisy observation. 
Although one-step denoising can greatly improve inference speed, it tends to produce overly smooth images and may lose fine details, which can lead to false alarms or mislead the subsequent detection stage. 
Consequently, we further introduce a frequency-guided perception loss that enforces consistency between the reconstructed image and the ground-truth background in the high-frequency domain. 
This constraint helps preserve background details such as edges and textures, while also strengthening the supervision signal in target regions.

Finally, the reconstructed object-free background image is concatenated with the original input image along the channel dimension and fed into the detection network. This dual-channel input provides explicit background prior information, enabling the detector to more accurately localize small targets by comparing the differences between the two images. Importantly, DiffIRSTD is designed as a plug-and-play framework that can be integrated with existing infrared small target detection models without increasing the computational complexity of the detection network, and it can be flexibly adapted to different deployment constraints.
\textbf{In summary, our contributions can be organized as follows,}
\vspace{-0.05in}
\begin{itemize}
\item We propose a novel diffusion-based framework for infrared small target detection, which offers a new perspective for denoising. We treat the small targets as a part of noise perturbation and use the diffusion reverse process to denoise them, thereby enhancing the contrast between the foreground and background.

\item To improve the efficiency of diffusion sampling, we present a one-step denoising strategy and additionally introduce a frequency-guided perceptual loss. The constraints from the high-frequency domain can effectively guide the denoising process to preserve the background details while focusing on the reconstruction quality of the target area.

\item Extensive experiments on three IRSTD benchmarks demonstrate that our proposed method outperforms state-of-the-art approaches, achieving superior detection performance.
\end{itemize}

\section{Related work}

\subsection{Infrared Small Target Detection}
In the field of deep learning-based IRSTD methods, the research utilizing CNNs has mainly focused on structural fine-tuning and loss function optimization to break through the bottleneck that small target features are prone to being submerged. DNA-Net~\cite{DNANet} proposes a dense nested interaction module to avoid small target loss during deep downsampling. UIU-Net~\cite{wu2022uiu} adopts a U-Net nested structure to strengthen multi-scale information retention. ISNet~\cite{zhang2022isnet} emphasizes the crucial role of target shape information by designing an edge block. MSHNet~\cite{liu2024infrared} proposes a scale and location-sensitive loss to improve the model's discriminative ability for targets of different scales and spatial positions. 
Transformer-based methods like SCTransNet~\cite{yuan2024sctransnet} deploy spatial-channel cross Transformer blocks at the skip connections of U-Net to bridge the semantic gap between the encoder and decoder. RKformer~\cite{zhang2022rkformer} introduces ViTs to balance local details and global dependencies. DATransNet~\cite{hu2025datransnet} proposes a DATrans to accurately extract target edge gradients. 
Recently, research on IRSTD has been extended to Mamba and SAM to enrich feature representations. MiM-ISTD~\cite{chen2024mim} proposes a "Mamba-in-Mamba" nested design, which achieves a leap in accuracy while maintaining linear computational overhead. IRMamba~\cite{zhang2025irmamba} introduces PDMamba to enhance target contour perception. In the SAM-based research, IRSAM~\cite{zhang2024irsam} is the first work to adapt SAM to the IRSTD task, utilizing WPMD and a GAD to finely reconstruct masks. Meanwhile, SAM-SPL~\cite{fu2025unified} leverages the general semantic knowledge of SAM to propose a unified self-prompt learning framework.

Different from the aforementioned architectures that focus on the explicit learning of target features, some low-rank decomposition studies~\cite{bai2010analysis,gao2013infrared,zhang2019infrared,wu2024rpcanet,zhang2025s} return to the physical essence of infrared imaging and 
formulate infrared small target detection as a background-target decoupling problem. Given an infrared image (\textbf{I}), the scene is typically modeled as,
\begin{equation}
    \textbf{I} = \textbf{B} + \textbf{T} + \textbf{N}
\end{equation}
\noindent where \textbf{B}, \textbf{T}, and \textbf{N} denote the background, target, and noise components, respectively. These methods estimate the background component via the deep model $\mathcal{F}(*)$ and subsequently detect targets through residual extraction,
\begin{equation}
\hat{\textbf{T}}=\textbf{I}-\hat{\textbf{B}} = \textbf{I}-\mathcal{F}(\textbf{I})
\label{eq:low-rank}
\end{equation}

For example, RPCANet~\cite{wu2024rpcanet} unfolds the iterative solution of Robust Principal Component Analysis (RPCA) into a network, combining low-rank priors with deep features to improve interpretability. LRRNet~\cite{zhang2025s} adopts a compression-reconstruction-subtraction paradigm, implicitly learning and generating low-rank background representations with structure-aware capabilities in a data-driven manner. 
Although LRRNet and our work both involve background estimation, the motivations and mechanisms are fundamentally different. LRRNet assumes that targets in the infrared image are sparse components over a low-rank background and performs target detection through low-rank decomposition and residual extraction. In contrast, our method formulates target removal as a generative denoising problem, where diffusion models can be leveraged to learn the underlying background distribution and generate a target-free image. The generated background is not used to directly locate targets but is instead provided as complementary contextual information to enhance target perception within the detector.
Briefly, LRRNet attempts to suppress the background, whereas our method attempts to understand the background.
Furthermore, our generated background serves as an auxiliary cue rather than a detection result, enabling the proposed framework to be seamlessly integrated into various existing detectors in a plug-and-play manner.

\subsection{Diffusion-based Object Detection Models}
As a class of generative models, Diffusion Models have demonstrated powerful capabilities in the field of computer vision. Their applications have transcended image generation and gradually expanded to discriminative tasks such as detection and segmentation. The pioneering work that first introduced diffusion models into object detection is DiffusionDet~\cite{chen2023diffusiondet}, which regards the generation process of target bounding boxes as a denoising process from a random noise distribution to real target boxes. DiffusionAD~\cite{zhang2025diffusionad} proposes a noise normalization paradigm and a single-step denoising mechanism for anomaly detection. It achieves unbiased reconstruction and accurate localization of abnormal regions through a dual-branch network.
Although both DiffusionAD and our method employ diffusion-based reconstruction, their underlying assumptions and objectives are fundamentally different. DiffusionAD assumes that anomalies lie outside the normal data distribution and reconstructs normal samples by removing semantic defects. In contrast, infrared small targets are not semantic anomalies but extremely weak signals embedded in cluttered backgrounds and sensor noise, often exhibiting substantial distribution overlap with the background. Consequently, our technical designs aim to efficiently purify the backgrounds to enhance target-background contrast rather than eliminating semantic abnormalities.

In recent years, diffusion models have been gradually introduced into IRSTD tasks, and relevant research has mainly developed along two paths: data augmentation and detection architecture. In the data augmentation paradigm, aiming at the lack of realism in traditional stitching methods, Diff-Mosaic~\cite{shi2024diff} designs a pixel coordination network combined with latent diffusion priors for resampling. Gaussian Agnostic Representation Learning~\cite{li2025exploiting} addresses the challenge of data scarcity by proposing a Gaussian-driven non-uniform quantization strategy and a two-stage diffusion generation framework. In the detection and feature modeling paradigm, DCFR-Net~\cite{fan2024diffusion} innovatively embeds implicit neural representation into conditional diffusion models, learning continuous high-resolution feature representations in a self-supervised manner. ISTD-diff~\cite{du2024istd} models IRSTD as a conditional mask generation task, directly outputting high-precision target masks through an iterative denoising process. 
Unlike these methods that use diffusion models to generate images with various foreground and background, 
our method treats targets as noise-like signals and utilizes diffusion models to remove them, thereby generating target-free background images for the subsequent detection. In addition, our reconstruction and detection networks are not tightly coupled but are independently separated. Therefore, it can be easily adapted to different detectors according to the requirements of deployment scenarios.

\section{Methodology}
\subsection{Overview}
\begin{figure*}[t]
    \centering  
    \includegraphics[width=0.95\linewidth]{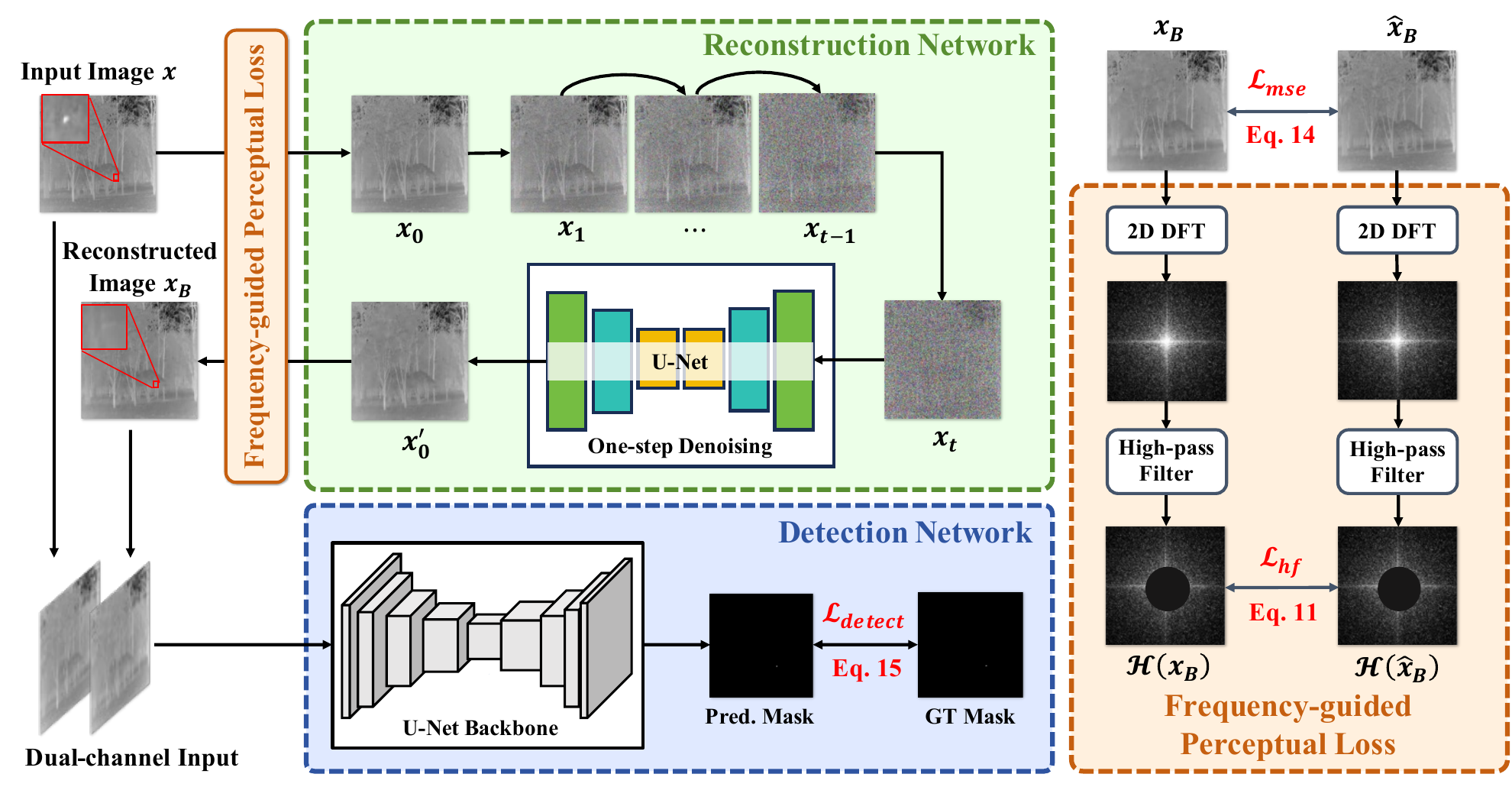}
    \vspace{-0.1in}
    \caption{\textbf{Schematic of our proposed DiffIRSTD, which consists of a reconstruction network and a detection network.} The reconstruction network is designed to generate a target-free background image $x_B$ from the input infrared image $x$. We further incorporate a one-step denoising strategy together with a frequency-guided perceptual loss. This design not only accelerates the sampling process but also preserves the denoising quality of the reconstructed background. The detection network takes as input a dual-channel representation formed by concatenating the original infrared image and the reconstructed image. By leveraging the explicit background prior, the network can effectively perceive target regions and predict the corresponding target mask.}
    \label{fig:framework}
    \vspace{-0.1in}
\end{figure*}

As aforesaid, the key idea of our DiffIRSTD is to enhance the contrast of small targets in infrared images using generative models, thereby improving the detector's ability to perceive them.
Unlike previous approaches that use diffusion models to either generate background-free target masks or as the detector to localize targets, we take a different perspective of treating small targets as noise-like perturbations in infrared images ($\textbf{T}\approx\textbf{N}$). Based on this assumption, we thus exploit the denoising property of diffusion models 
to learn the conditional distribution of target-free backgrounds,
\begin{equation}
    \textbf{B}_{\mathrm{clean}} \sim p(\textbf{B}|\textbf{I})
\end{equation}
\noindent where $\textbf{B}_{\mathrm{clean}}$ denotes the generated target-free background conditioned on the observed image (\textbf{I}).
The generated background is not used for residual computation, but is instead provided as contextual information to a detector (compared with Eq.~\ref{eq:low-rank}),
\begin{equation}
    \hat{\textbf{T}}=\mathcal{D}(\textbf{I},\textbf{B}_{\mathrm{clean}})
\end{equation}
\noindent where $\mathcal{D}(\cdot)$ denotes the target detector.
Consequently, existing methods regard the background as interference to be suppressed, whereas our method explicitly models the background distribution and exploits the generated background as informative scene context. By comparing the original image with the reconstructed target-free image, the detector can more easily identify the location and precise boundaries of the target.
Figure~\ref{fig:framework} illustrates the overall architecture of our proposed diffusion-based framework for infrared small target detection.

In the remainder of this section, we first explain the basic principles of diffusion models, including the diffusion and denoising processes (in Sec.~\ref{subsec:preliminary}). Next, we describe how diffusion models are utilized to denoise infrared images in Sec.~\ref{subsec:reconstruction}, focusing on our one-step denoising strategy and the proposed frequency-guided perceptual loss, which together improve both the efficiency and quality of reconstructing background images without targets. We then discuss how to leverage the generated target-free infrared images to strengthen the detector's ability to perceive small targets (see Sec.~\ref{subsec:detection}). Subsequently, we present the workflow of obtaining synthetic background images using morphological operations, which serve as pseudo-labels for training the diffusion model (in Sec.~\ref{subsec:label}). Finally, we detail the training and inference procedures of the proposed DiffIRSTD framework in Sec.~\ref{subsec:training}.

\subsection{Preliminary \label{subsec:preliminary}}
Denoising Diffusion Probabilistic Models (DDPM) \cite{ho2020denoising} are a class of latent-variable generative models that define the data generation process through two Markov chains: a forward diffusion process and a reverse denoising process.

\noindent \textbf{Forward Diffusion Process.} In the forward process, Gaussian noise is gradually added to the data. Given a real data sample $x_0 \sim q(x_0)$, a Markov chain is defined in which noise is incrementally injected over $T$ time steps,
\begin{equation}
q(x_t|x_{t-1}) = \mathcal{N}(x_t; \sqrt{1-\beta_t}x_{t-1}, \beta_t\mathbf{I})
\end{equation}
where $\beta_t \in (0, 1)$ denotes a predefined noise schedule that controls the noise level at each step. By applying the reparameterization trick, a noisy sample at any timestep $t$ can be directly obtained from the original data $x_0$,
\begin{equation}
x_t = \sqrt{\bar{\alpha}_t}x_0 + \sqrt{1-\bar{\alpha}_t}\epsilon
\label{eq:diffusion}
\end{equation}
where $\epsilon \sim \mathcal{N}(0,\mathbf{I})$, $\alpha_t = 1-\beta_t$, and $\bar{\alpha}_t = \prod_{i=1}^{t} \alpha_i$.

\noindent \textbf{Reverse Denoising Process.} The reverse process aims to recover clean data from pure Gaussian noise $x_T\sim\mathcal{N}(0, \mathbf{I})$ through iterative denoising. A parameterized model $p_\theta(x_{t-1}|x_t)$ is trained to approximate the posterior distribution,
\begin{equation}
p_\theta(x_{t-1}|x_t) = \mathcal{N}(x_{t-1}; \mu_\theta(x_t,t), \Sigma_\theta(x_t,t))
\end{equation}
In practice, the variance is often fixed as $\Sigma_\theta(x_t,t) = \sigma_t^2\mathbf{I}$, and a neural network $\epsilon_\theta(x_t,t)$ is trained to predict the noise component $\epsilon$ added at step $t$. The denoising step from $x_{t}$ to $x_{t-1}$ can be computed directly as,
\begin{equation}
x_{t-1} = \frac{1}{\sqrt{\alpha_{t}}}(x_t - \frac{1 - \alpha_{t}}{\sqrt{1 - \bar{\alpha}_{t}}}\epsilon_\theta(x_t,t)) + \sigma_t z
\label{eq:denoise}
\end{equation}
where $z \sim \mathcal{N}(0, \mathbf{I})$ is standard Gaussian noise. The training objective of diffusion models is to minimize the mean squared error between the predicted and ground-truth noise,
\begin{equation}
\mathcal{L}_{mse} = \mathbb{E}_{t,x_0,\epsilon}\left[\|\epsilon - \epsilon_\theta(x_t,t)\|^2\right]
\label{eq:noise_loss}
\end{equation}

\subsection{Reconstruction Network \label{subsec:reconstruction}}
Revisiting the imaging characteristics of infrared small targets, we observe that although infrared targets are small and lack distinct structural features, they often exhibit subtle local intensity deviations from their surrounding background. These deviations may be visually difficult to perceive due to complex scenes, illumination conditions, and sensor noise. From a statistical perspective, infrared small targets can therefore be regarded as noise-like signals relative to the smooth background. Generating a target-free image is thus analogous to the denoising process of a diffusion model, which progressively removes noise (\textit{e.g.}, target) to recover a clean signal (\textit{e.g.}, background). Therefore, we adopt a diffusion model as the reconstruction network to generate the background image. 

Given an infrared image $x \in \mathbb{R}^{H \times W}$ containing small targets, we first apply Eq.~\ref{eq:diffusion} to gradually add Gaussian noise, so that after $t$ steps, the image becomes increasingly corrupted and the targets are overwhelmed by noise. The resulting noisy image is then fed into the reverse diffusion process, which iteratively denoises the observation according to Eq.~\ref{eq:denoise}. During this reverse process, the model is encouraged to reconstruct a clean background image while suppressing target-related ``noise''.
As a result, we obtain a reconstructed image $x_B \in \mathbb{R}^{H \times W}$ that approximates the scene without targets.

However, according to Eq.~\ref{eq:denoise}, the reverse diffusion process progressively denoises the perturbed image from $x_t$ to $x_0$ through multiple iterative steps. Each iteration corresponds to a single denoising transition, and performing all $t$ steps during inference leads to substantial computational overhead. This iterative sampling procedure significantly increases the inference time and limits the practicality of diffusion models in real-time infrared small target detection. Is it possible to directly predict the background image $x_0$ from a noisy observation $x_t$ in a single step?

\noindent \textbf{One-step Denoising Strategy.} 
During the theoretical derivation of diffusion models \cite{ho2020denoising}, it has been proven that the clean image $x_0$ can be estimated directly from a noisy sample $x_t$ by rearranging the forward formulation Eq.~\ref{eq:denoise}. However, this direct estimation generally produces images of lower quality compared to multi-step iterative sampling. The primary reason is that when $x_t$ approaches pure Gaussian noise, the prediction errors in $\epsilon$ may be amplified, leading to distortions and artifacts. Differently, our purpose is not to synthesize images from random noise, but to remove small target ``noise''. Since the background of infrared images is typically smooth, recovering a target-free background image from a noisy observation is easier than reconstructing a natural image from pure noise. Therefore, given an input image $x$ and its noisy version $x_t$ obtained at diffusion step $t$, we directly employ the one-step denoising process to estimate the background image, 
\begin{equation}
x_0 = \frac{1}{\sqrt{\bar{\alpha}_t}}(x_{t} - \sqrt{1-\bar{\alpha}_t}\,\epsilon_\theta(x_t,t)).
\label{eq:one_step}
\end{equation}
where $x_B = x_0$. By skipping intermediate iterative steps, the one-step denoising strategy substantially accelerates inference and reduces computational cost. Intuitively, if $t$ is too large, one-step denoising encounters the above limitation again and may degrade image quality. Conversely, a small $t$ makes the noise-like target signals difficult to remove by one step.
Therefore, selecting an appropriate noise level balances the target denoising and background fidelity, which will be discussed in the experimental section.

\noindent \textbf{Frequency-guided Perceptual Loss.}
Infrared small targets typically occupy only a tiny fraction of the image area, which leads to highly imbalanced supervision during the reconstruction process. Moreover, even minor reconstruction errors in the background may introduce spurious high-frequency artifacts, which can mislead the subsequent detection network. These high-frequency components, especially edges and textures, provide important cues for distinguishing true targets from complex background clutter.

To explicitly preserve these structural details, we introduce a high-frequency perceptual constraint in the frequency domain. Specifically, the predicted background image $x_B$ and the ground-truth background $\hat{x}_B$ are first transformed into the frequency domain using two-dimensional discrete Fourier transform (2D DFT),
\begin{equation}
\mathcal{F}(X)(u, v) = \sum_{m=0}^{H-1} \sum_{n=0}^{W-1} X(m, n) e^{-j2\pi \left( \frac{um}{H} + \frac{vn}{W} \right)}
\end{equation}
where $X \in [x_B, \hat{x}_B]$, and $(u, v)$ denote the frequency-domain coordinates. The zero-frequency component is then shifted to the center of the spectrum to obtain $\mathcal{F}_{\text{shift}}(X)$. 
To further reduce the dynamic range of spectral magnitudes, we compute the logarithmic amplitude spectrum and normalize it spatially,
\begin{equation}
\hat{\mathcal{F}}_{\text{shift}}(X) = \text{Norm} \left[ \text{log} \left( 1+|\mathcal{F}_{\text{shift}}(X)| \right) \right]
\end{equation}
Next, a binary high-pass mask $M(u,v) \in \{0,1\}^{H \times W}$ is constructed to retain only frequency components whose radial distance from the spectrum center exceeds a predefined cutoff frequency $r_c$.
Thereby, the high-frequency representation $\mathcal{H}(X)$ can be obtained by,
\begin{equation}
M(u,v)=
\begin{cases}
1, & \sqrt{(u-u_c)^2 + (v-v_c)^2} > r_c \\
0, & \text{otherwise}
\end{cases}
\end{equation}
\begin{equation}
\mathcal{H}(X) = \hat{\mathcal{F}}_{\text{shift}}(X) \cdot M
\end{equation}
where $(u_c, v_c)$ means the coordinates of spectrum center. Finally, the high-frequency perceptual loss is defined as the mean squared error between the masked spectra of the reconstructed and ground-truth backgrounds,
\begin{equation}
\mathcal{L}_{hf} = \|\mathcal{H}(x_B)-\mathcal{H}(\hat{x}_B)\|^2
\label{eq:frequency}
\end{equation}

This frequency-guided perceptual loss encourages the reconstruction network to pay closer attention to high-frequency discrepancies introduced by target regions while simultaneously preserving intrinsic background structures during one-step denoising. As a result, the model is able to suppress target-induced noise without sacrificing the sharpness and texture of the reconstructed background, leading to higher-quality target-free images.

\subsection{Detection Network \label{subsec:detection}}
Similar to previous approaches~\cite{ren2021dnanet,wu2022uiu,yuan2024sctransnet,zhang2025irmamba}, the objective of our detection network is to accurately predict the location mask of small targets. But, differently, benefiting from the background prior provided by our reconstruction network, the detection network can additionally receive the corresponding target-free background image, so as to highlight target regions through comparison analysis.

Specifically, we first take as input the input infrared image $x \in \mathbb{R}^{H\times W}$ and the reconstructed background image $x_B \in \mathbb{R}^{H\times W}$, and concatenate them along the channel dimension to form a dual-channel input,
\begin{equation}
    x_{dual} = \text{Concat}(x, x_B) \in \mathbb{R}^{2 \times H \times W}.
\end{equation}
This dual-channel representation is thus fed into the detection network to localize target regions by learning the discrepancy between the two channels. 

The detection network follows a standard U-Net architecture~\cite{ronneberger2015u}, consisting of an encoder, a decoder, and skip connections. The encoder extracts multi-scale semantic features through progressive down-sampling, while the decoder gradually restores spatial resolution via up-sampling. Skip connections fuse features from corresponding encoder and decoder stages, helping preserve fine-grained spatial details that are important for small target localization.

To further enhance feature discriminability, we adopt the channel attention module, which is inserted after each encoder block. The attention module aggregates contextual information along spatial and channel dimensions, and then calculates the channel weights to rescale the importance of feature maps. Formally, given an input feature map $f^{in} \in \mathbb{R}^{c \times h \times w}$, the attention-enhanced output is computed as,
\begin{equation}
f^\text{out} = f^\text{in} \cdot \sigma\left(\text{MLP}(\text{GMP}(f^\text{in}) + \text{GAP}(f^\text{in})  )\right)
\end{equation}
where $\sigma$ is the sigmoid activation, MLP means Multilayer Perceptron, and GMP and GAP represent global max pooling and global average pooling, respectively. In this way, the network can adaptively emphasize channels that are more relevant to small targets while suppressing those dominated by background interference. The refined feature maps are then forwarded to the next encoder stage and to the corresponding skip connections. The decoder finally gradually interprets and outputs the predicted target mask.

\noindent \textbf{Discussion.} 
In this paper, we intentionally adopt a simple U-Net as the baseline detection network. This design choice allows us to more clearly evaluate the effectiveness of the proposed reconstruction network and directly support our motivation. Importantly, our DiffIRSTD framework does not depend on any specific detector architecture.
The reconstruction network is designed as a plug-and-play component that generates a clean background image. This reconstructed background can be concatenated with the original infrared image to form a dual-channel representation, which can be readily used by existing infrared small target detection models. As shown in Tab.~\ref{tab:different detector}, integrating the reconstruction network with various state-of-the-art detectors consistently improves their performance.
Furthermore, different from previous diffusion-based approaches~\cite{li2024dense,rong2026novel,chenhao2025diffusion}, the reconstruction and detection networks in our framework are loosely connected. This decoupled design provides greater flexibility in practical deployment. Depending on computational constraints, the detection network can be replaced with either lighter or more powerful models without modifying the reconstruction network. Alternatively, the reconstruction network can be deployed independently on the cloud as a preprocessing step during inference.

\subsection{Background Acquisition Strategy \label{subsec:label}}

\begin{figure}[t]
    \centering  
    \includegraphics[width=0.95\linewidth]{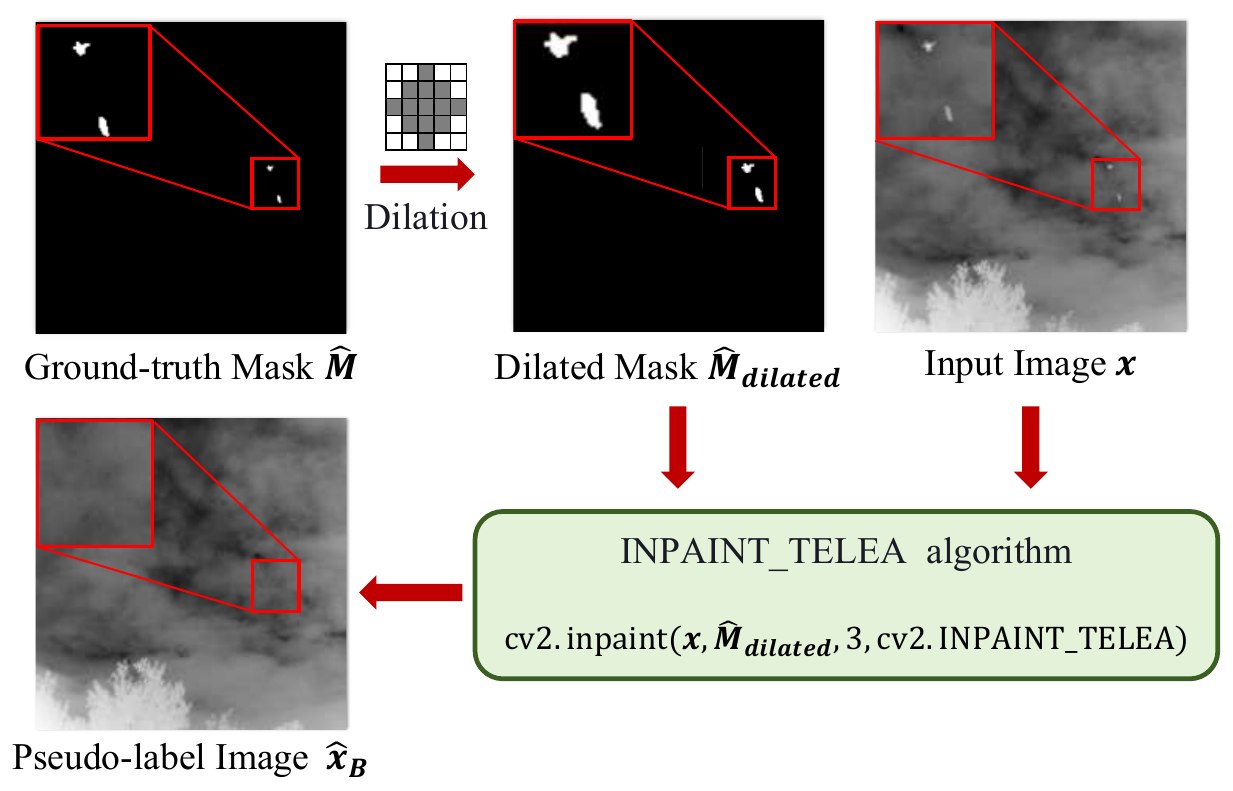}
    \vspace{-0.1in}
    \caption{\textbf{The workflow of acquiring the pseudo-label target-free background images.} Best viewed in color and zoom in.}
    \label{fig:inpaint}
    \vspace{-0.1in}
\end{figure}

Existing infrared small target datasets provide only infrared images and their corresponding target masks, while ground-truth background images are not available. However, such background images are necessary to supervise the reconstruction process described in Sec.~\ref{subsec:reconstruction}. 
To address this issue, we propose an inpainting-based background acquisition strategy that removes target regions using the provided masks and fills them to synthesize high-quality background images for training. The workflow of this strategy is illustrated in Fig.~\ref{fig:inpaint}.

More specifically, given an infrared image $x$ and its target mask $\hat{M}$, we first apply a threshold of $0.5$ to ensure that the mask is strictly binary. Next, a morphological dilation operation with a 5$\times$5 kernel is applied to the mask to slightly enlarge the target region. This step ensures that the dilated mask $\hat{M}_{dilated}$ fully covers the target boundaries and nearby pixels, preventing residual target artifacts after inpainting.
Finally, we employ the OpenCV implementation of the Telea inpainting algorithm~\cite{article} to fill the masked regions, with the inpainting radius set to 3 pixels. The resulting image is thus used as the pseudo-label background image, denoted by $\hat{x}_B$. This algorithm preserves structural continuity and texture consistency in the surrounding background, making it suitable for the reconstruction of small regions.

\subsection{Training and Inference \label{subsec:training}}
\noindent \textbf{Training.}
Our framework is trained in an end-to-end manner. For the reconstruction network, it learns to generate high-quality backgrounds under the joint supervisions of the pixel-level reconstruction loss and frequency-guided perceptual loss, 
\begin{equation}
    \mathcal{L}_{recon}=\mathcal{L}_{mse}(x_B, \hat{x}_B)+\lambda\cdot\mathcal{L}_{hf}(x_B, \hat{x}_B)
\label{eq:rec_loss}
\end{equation}
where $\lambda$ is the balancing coefficient to control the importance of the frequency-guided perceptual loss. $\mathcal{L}_{mse}$ denotes the mean squared error between the predicted and ground-truth images, and we empirically find it is better than the noise supervision in Eq.~\ref{eq:noise_loss}. Here, the high-frequency loss $\mathcal{L}_{hf}$ plays a more similar role to that of the classifier-based guidance, guiding the denoising process to generate high-quality target-free background images.

Meanwhile, the detection network learns to locate targets and output segmentation masks under the supervision of SoftIoU Loss and BCE Loss,
\begin{equation}
    \mathcal{L}_{detect}=\mathcal{L}_{softiou}(M,\hat{M})+\mathcal{L}_{bce}(M,\hat{M})
\end{equation}
where $M$ and $\hat{M}$ are the predicted and ground-truth masks. The SoftIoU loss $\mathcal{L}_{softiou}$ optimizes the intersection over union between target regions, while the BCE loss $\mathcal{L}_{bce}$ ensures the pixel-level classification accuracy. The collaboration between two functions improves the quality of target localization.

\noindent \textbf{Inference.} 
During inference, the input image $x$ first passes through the reconstruction network to get the denoised target-free image $x_{B}$, and then both of them are fed into the detection network to predict the target mask $M$.

\section{Experiments}

\begin{table*}[t]
  \centering
  \small
    \caption{\textbf{Comparison with existing IRSTD approaches on IRSTD-1K, NUDT-SIRST, and NUAA-SIRST datasets.} The evaluation metrics are IoU ($10^{-2}$), $P_d$ ($10^{-2}$), and $F_a$ ($10^{-6}$). The first and second best results are highlighted in \textbf{bold} and \underline{underlined}, respectively. $\dagger$ denotes methods that use diffusion models for either data augmentation, feature learning, or diffusion-based detectors.}
    \setlength{\tabcolsep}{3.3mm}{
    \begin{tabular}{lcccccccccc}
    \toprule
    \multicolumn{1}{c}{\multirow{2}{*}{\textsc{MethodS}}} & \multirow{2}{*}{\textsc{Backbone}} & \multicolumn{3}{c}{\textbf{IRSTD-1K}} & \multicolumn{3}{c}{\textbf{NUDT-SIRST}} & \multicolumn{3}{c}{\textbf{NUAA-SIRST}} \\
    \cmidrule(r){3-5} \cmidrule(r){6-8} \cmidrule(r){9-11}
    & & IoU $\uparrow$ & P$_{d} \uparrow$ & F$_{a} \downarrow$ & IoU $\uparrow$ & P$_{d} \uparrow$ & F$_{a} \downarrow$ & IoU $\uparrow$ & P$_{d} \uparrow$ & F$_{a} \downarrow$ \\
    \midrule
    Top-Hat~\cite{bai2010analysis}  & Traditional & 10.06 & 75.11 & 1432 & 20.72 & 78.41 & 166.7 & 7.143 & 79.84 & 1012 \\
    IPI~\cite{gao2013infrared}  & Traditional & 27.92 & 81.37 & 16.18 & 17.76 & 74.49 & 41.23 & 25.67 & 84.63 & 16.67 \\
    PSTNN~\cite{zhang2019infrared}  & Traditional & 24.57 & 71.99 & 35.26 & 14.85 & 66.13 & 44.17 & 30.30 & 72.80 & 48.99 \\
    MSLSTIPT~\cite{9203993}  & Traditional & 11.43 & 79.03 & 1524 & 8.342 & 47.40 & 888.1 & 10.30 & 82.13 & 1131 \\
    \midrule
    MTU-Net~\cite{wu2023mtu}  & CNN + Trans. & 66.11 & 93.27 & 36.80 & 74.85 & 93.97 & 46.95 & 74.78 & 93.54 & 22.36 \\
    SCTransNet~\cite{yuan2024sctransnet} & CNN + Trans. & 68.03 & 93.27 & 10.74 & 94.09 & 98.62 & 4.290 & 77.50 & 96.95 & 13.92 \\
    HSTNet~\cite{li2025hstnet} & CNN + Trans. & 69.79 & 91.16 & 10.70 & - & - & - & 78.14 & \textbf{100} & 14.72 \\
    ISTD-Diff$^\dagger$~\cite{du2024istd} & Diffusion & - & - & - & - & - & - & 76.09 & 97.25 & 11.91 \\
    DCDNet$^\dagger$~\cite{li2024dense} & Diffusion & - & - & - & 90.34 & 98.62 & - & - & - & - \\
    IRMamba~\cite{zhang2025irmamba} & Mamba & \underline{70.04} & \textbf{95.81} & \textbf{5.920} & \underline{95.18} & 99.26 & \textbf{1.309} & \underline{80.20} & \underline{99.36} & \textbf{0.887} \\
    \midrule
    FC3-Net~\cite{zhang2022exploring} & CNN & 65.07 & 91.54 & 15.55 & 78.56 & 93.86 & 23.92 & 72.44 & 98.14 & 10.85 \\
    DNA-Net~\cite{li2022dense} & CNN & 65.90 & 90.91 & 12.24 & 88.19 & 98.83 & 9.000 & 75.80 & 95.82 & 8.780 \\
    UIU-Net~\cite{wu2022uiu} & CNN & 66.15 & 93.98 & 22.07 & 93.48 & 98.31 & 7.790 & 76.91 & 95.82 & 14.13 \\
    ISTDU~\cite{9674870} & CNN & 66.36 & 93.60 & 53.10 & 89.55 & 97.67 & 13.44 & 75.52 & 96.58 & 14.54 \\
    AGPCNet~\cite{zhang2021agpcnet} & CNN & 66.29 & 92.83 & 13.12 & 88.87 & 97.20 & 10.02 & 75.69 & 96.48 & 14.99 \\
    Dim2Clear~\cite{10091167} & CNN & 66.34 & 93.75 & 20.93 & 81.37 & 96.23 & 9.170 & 77.29 & 99.10 & 6.72 \\
    MMLNet~\cite{li2025multi} & CNN & 67.21 & \underline{94.28} & 14.00 & 81.81 & 98.43 & 11.77 & 78.71 & 98.88 & 25.71 \\
    PConv\cite{yang2025pinwheel} & CNN & 67.93 & 94.22 & \underline{9.19} & - & - & - & - & - & - \\
    L2SKNet~\cite{wu2024saliency} & CNN & 67.81 & 90.24 & 17.46 & 93.58 & 97.57 & 5.330 & 73.43 & 98.17 &  20.82 \\
    Diff-Mosaic$^\dagger$~\cite{shi2024diff} & CNN & - & - & - & 91.18 & \underline{99.47} & 1.92 & - & - & - \\
    DCFRNet$^\dagger$~\cite{fan2024diffusion} & CNN & 65.41 & 93.60 & 7.340 & 86.93 & 98.26 & 2.48 & 76.23 & 99.08 & \underline{6.52} \\
    FAA-Net~\cite{zhuang2025faa} & CNN & 68.63 & 93.88 & 14.65 & 93.53 & 99.15 & 2.37 & 77.57 & \textbf{100} & 17.39 \\
    \midrule
    \rowcolor{gray!10} Upper Bound & - & 94.21 & 98.99 & 0.000 & 97.20 & 100.0 & 0.48 & 85.96 & 99.84 & 3.11 \\
    U-Net Baseline & CNN & 67.13 & 93.93 & 26.64 & 93.97 & 99.04 & 3.44 & 78.35 & 94.29 & 15.36 \\
    DiffIRSTD (\textit{Ours}) & CNN & \textbf{72.13} & 91.91 & 13.64 & \textbf{95.91} & \textbf{99.49} & \underline{1.72} & \textbf{80.73} & 96.95 & 7.55 \\
    \bottomrule
    \end{tabular}
    }
  \label{tab:comparsion_SOTA}
\end{table*}

\subsection{Experimental Setups}
\noindent \textbf{Datasets.} Three public datasets of IRSTD-1K~\cite{ISNet}, NUDT-SIRST~\cite{DNANet}, and NUAA-SIRST~\cite{ACMNet} are used for evaluation. Following common practice in prior work~\cite{ACMNet,DNANet}, we split both NUDT-SIRST and NUAA-SIRST evenly into training and testing sets (5:5). For IRSTD-1K, we use 80\% of the samples for training and the remaining for testing (8:2), same as the official protocol for fair comparisons.

\noindent \textbf{Implementation Details.} Our framework is implemented on the PyTorch platform using one NVIDIA GeForce RTX 4090 GPU.
The reconstruction network is based on the U-Net architecture \cite{dhariwal2021diffusion,zhang2025diffusionad}. During training, all input images are resized to $256\times256$ with the augmentation of random cropping, padding, and normalization. Both reconstruction and detection models are jointly trained with a batch size of $8$ for a total of $800$ epochs. We use the Adam optimizer with an initial learning rate of $1 \times 10^{-4}$, and the learning rate is adjusted using a cosine annealing schedule. 
We evaluate models using three widely-used metrics to provide a balanced assessment, where Intersection over Union ($\text{IoU} \times 10^{-2}$) evaluates geometric fidelity, detection probability ($F_d \times 10^{-2}$) ensures high recall of targets, and false-alarm rate ($F_a \times 10^{-6}$) penalizes false activations. These indicators jointly quantify localization fidelity, recall, and specificity, offering a holistic assessment of detection accuracy, precision, and reliability.

\begin{figure*}[t]
    \centering  
    \includegraphics[width=1\linewidth]{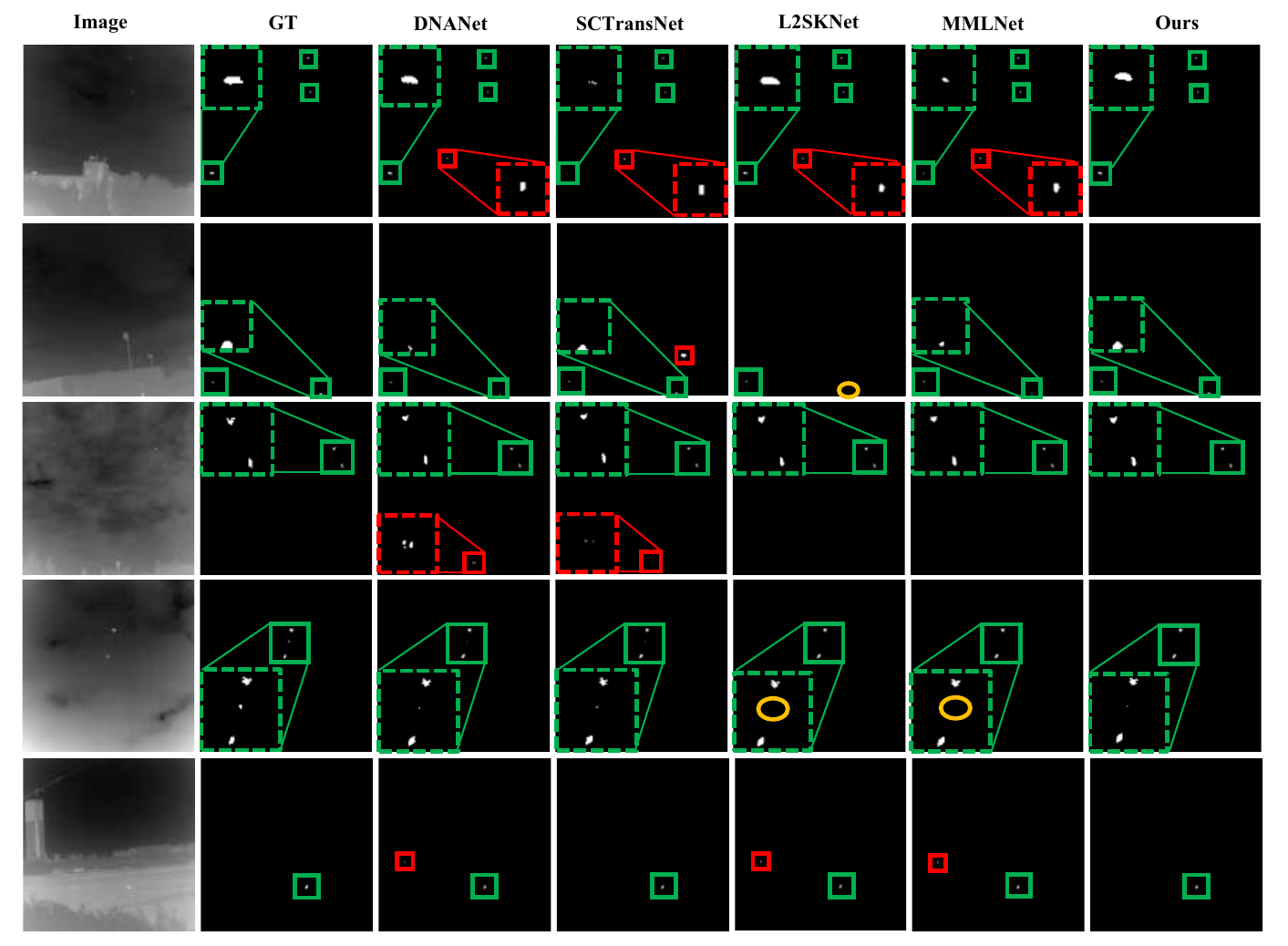}
    \caption{\textbf{Visualization of different IRSTD methods.} The green boxes indicate correctly detected targets, red boxes denote false alarms, and orange circles represent missed targets. Additionally, dashed boxes show close-up views of the targets for clearer visualization.}
    \label{fig:comparison}
\end{figure*}

\subsection{Comparisons with the State-of-the-arts}
We first compare our method with previous state-of-the-art approaches on the IRSTD-1K, NUDT-SIRST, and NUAA-SIRST datasets. All results of these competitors are taken from the reported results in the published papers.
As shown in Tab.~\ref{tab:comparsion_SOTA}, our DiffIRSTD achieves consistent improvements across all three datasets, demonstrating its effectiveness for infrared small target detection. Specifically, \textbf{(1)} our method attains the best IoU scores, reaching 72.13 on IRSTD-1K, 95.91 on NUDT-SIRST, and 80.73 on NUAA-SIRST, indicating improved localization accuracy. In terms of false-alarm rate ($F_a$), our method achieves competitive performance, particularly on NUDT-SIRST, while maintaining a reasonable balance between detection accuracy and false positives across datasets.
\textbf{(2)} Compared with existing diffusion-based approaches (\textit{e.g.}, ISTD-Diff, DCDNet, Diff-Mosaic, and DCFRNet), we beat them by a clear margin across datasets, suggesting that the proposed reconstruction strategy provides more effective background priors for detection.
\textbf{(3)} In addition, compared with hybrid backbone methods such as IRMamba and HSTNet, our method also achieves slightly higher IoU on all three datasets. Although IRMamba reports lower $F_a$ in some cases, we offer a competitive trade-off between localization accuracy and false-alarm suppression, despite using a relatively simple CNN-based detector.
\textbf{(4)} The ``Upper Bound'' results are obtained by bypassing the generation module and directly training the detector with the pseudo-label background image $\hat{x}_B$. This setting provides an estimate of the maximum achievable performance of the detection stage under ideal background reconstruction. As shown in Tab.~\ref{tab:comparsion_SOTA}, the upper bound substantially outperforms existing methods and achieves remarkably strong results on IRSTD-1K and NUDT-SIRST. 
This observation strongly validates the effectiveness of our motivation and the quality of our background acquisition strategy.
Moreover, the remaining gap between DiffIRSTD and the upper bound suggests considerable room for improvement, indicating that more accurate target-free background generation could further enhance detection performance.

Figure~\ref{fig:comparison} presents visualization comparisons of the inference results between our method and other state-of-the-art approaches, such as DNANet~\cite{DNANet}, SCTransNet~\cite{yuan2024sctransnet}, L2SKNet~\cite{wu2024saliency}, MMLNet~\cite{li2025multi}. 
As illustrated by the objects highlighted with green bounding boxes in the first and second rows, the predictions generated by our method exhibit superior alignment with the ground truth, demonstrating its enhanced capacity for precise boundary delineation. Furthermore, across all five visualized samples, the comparative methods are prone to false alarms and missed detections, whereas our approach effectively suppresses these prevalent segmentation errors.

\subsection{Ablation Study}
Next, we provide in-depth studies to dissect the efficacy of our designs. 
Without loss of generality, we conduct ablation studies on IRSTD-1K and use the same hyperparameters for other datasets. More quantitative analysis and qualitative results can be found in the supplementary materials.

\begin{table}[t]
  \centering
  \caption{Ablation study on key components of DiffIRSTD.}
  \small
  \setlength{\tabcolsep}{0.6mm}
  \begin{tabular}{cccccc}
  \toprule
  \textbf{Detection} & \textbf{Reconstruction} & \textbf{Frequency-guided}  & 
  \multicolumn{3}{c}{\textbf{IRSTD-1K}} \\
  \cline{4-6}
  \textbf{Network} & \textbf{Network} & \textbf{Perceptual Loss} & IoU $\uparrow$ & $P_d \uparrow$ & $F_a \downarrow$ \\
  \midrule
  \checkmark & & & 67.13 & 93.93 & 26.64 \\ 
  \checkmark & \checkmark & & 69.99 & 92.92 & 25.92 \\ 
  \checkmark & \checkmark & \checkmark & 72.13 & 91.91 & 13.64 \\
  \bottomrule
  \end{tabular}
  \label{tab:ablation integrity}
\end{table}

\begin{table}[t]
  \centering
  \caption{Ablation study of reconstruction loss on IRSTD-1K.}
  \small
  \setlength{\tabcolsep}{2.7mm}{
  \begin{tabular}{cccc}
  \toprule
  \textsc{loss variants} & IoU $\uparrow$ & P$_d \uparrow$ & F$_a \downarrow$ \\
  \midrule
  $\mathcal{L}_{mse}(\epsilon_\theta,\epsilon)$ (Eq.~\ref{eq:noise_loss}) & 67.14 & 91.24 & 32.05 \\
  $\mathcal{L}_{mse}(x_B, \hat{x}_B)$ (in Eq.~\ref{eq:rec_loss}) & 69.99 & 92.92 & 25.92 \\
  $\mathcal{L}_{mse}(x_B, \hat{x}_B) + \mathcal{L}_{hf}$ (Eq.~\ref{eq:rec_loss}) & 72.13 & 91.91 & 13.64 \\
  \bottomrule
  \end{tabular}}
  \label{tab:ablation loss design}
\end{table}

\begin{table}[t]
  \centering
  \caption{Ablation study of diffusion step $t$ in Eq.~\ref{eq:one_step}.}
  \small
  \setlength{\tabcolsep}{6mm}{
  \begin{tabular}{cccc}
  \toprule
  \textsc{time steps} & IoU $\uparrow$ & P$_d \uparrow$ & F$_a \downarrow$ \\
  \midrule
  30 & 72.16 & 91.58 & 18.95 \\
  50 & 72.13 & 91.91 & 13.64 \\
  70 & 70.69 & 91.58 & 12.63 \\
  90 & 71.27 & 93.93 & 14.45 \\
  120 & 68.94 & 90.90 & 16.13 \\
  \bottomrule
  \end{tabular}}
  \label{tab:ablation t}
\end{table}

\noindent \textbf{Efficacy of Designs in Reconstruction Network.}
Table~\ref{tab:ablation integrity} presents the results with different components. 
Our baseline is to solely activate the detection network with U-Net architecture. It obtains an IoU of 67.13 with a relatively high false-alarm rate of 26.64.
After incorporating with the reconstruction network, the IoU increases to 69.99, indicating that the reconstructed background provides useful prior information for enhancing target localization. Meanwhile, a slight reduction in $F_a$ is also observed, suggesting improved suppression of background interference.
Finally, we attach the proposed frequency-guided perceptual loss, and the performance is significantly improved. It strongly demonstrates the efficacy of our design, allowing the detector to locate target boundaries more accurately while avoiding false detections caused by blurred backgrounds.

We also investigate in Tab.~\ref{tab:ablation loss design} the impact of different reconstruction losses on the detection performance. Specifically, we observe that replacing the standard noise supervision of the diffusion model with image supervision can effectively improve the performance. We explain that small targets are treated as slight noise perturbations in the background images, so the image supervision offers a smoothing loss bias, and it is task‑aligned with the subsequent detection network.

\begin{figure*}[t]
    \centering  
    \includegraphics[width=1\linewidth]{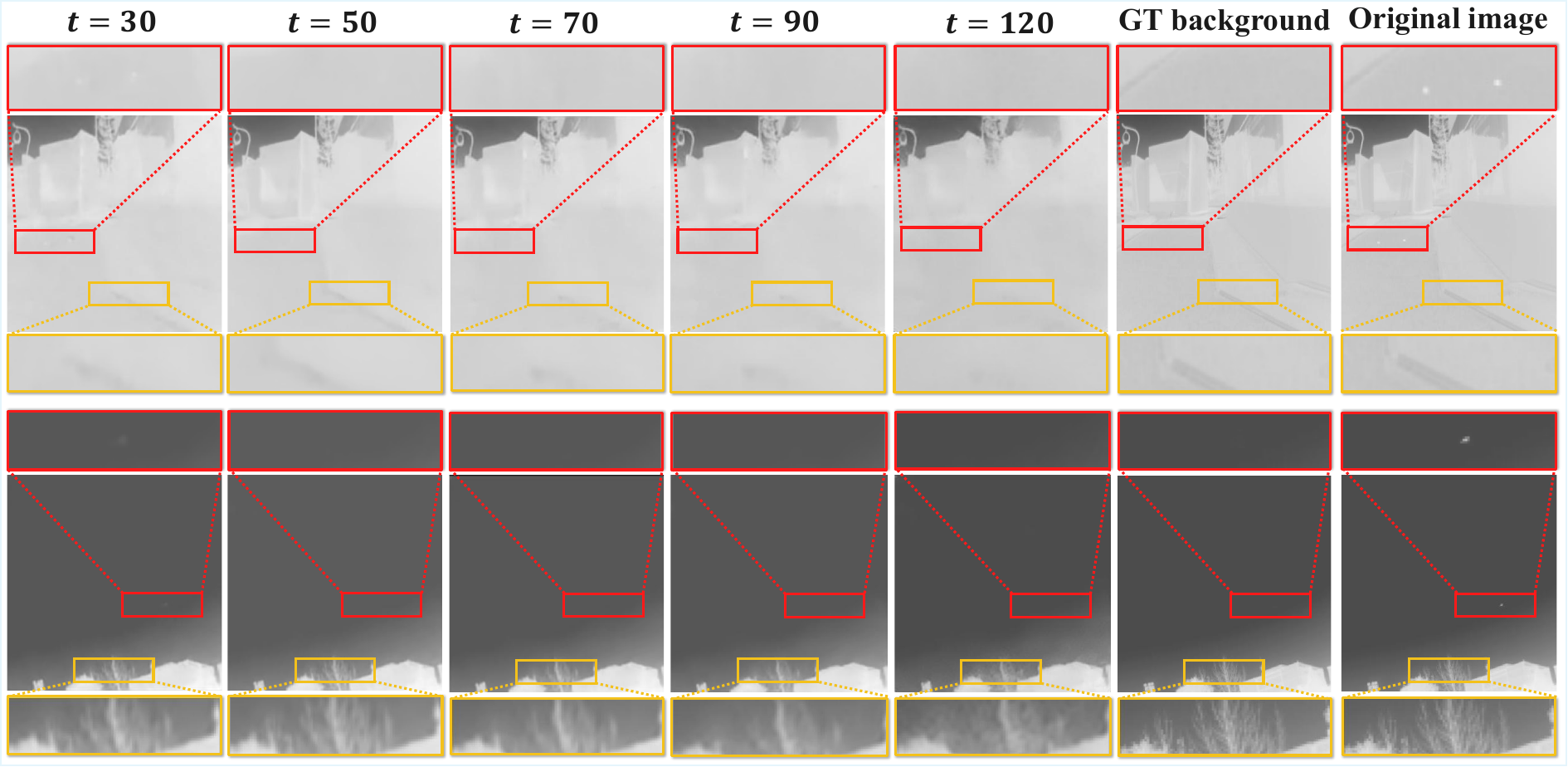}
    \caption{\textbf{Visualizations of reconstructed background images with different diffusion time steps $t$.} The red and yellow boxes denote close-ups of the foreground target and background details, respectively. Best viewed zoomed in.}
    \label{fig:t vis}
\end{figure*}

\noindent \textbf{Analysis of Diffusion Steps $t$.}
As discussed in Sec.~\ref{subsec:reconstruction}, the diffusion step $t$ plays an important role in reconstruction quality. We therefore analyze the effect of different time steps in Tab.~\ref{tab:ablation t}. Two main observations can be drawn: \textbf{(1)} within the range of $30 \sim 90$ steps, the overall performance variation is relatively small; \textbf{(2)} moderate step values (\textit{e.g.}, $t=50$ or $70$) tend to yield better results. In addition, the qualitative results in Fig.~\ref{fig:t vis} show that as $t$ increases, the removal of targets does not improve significantly, while the background details become progressively blurred. This is because introducing excessive noise makes one-step denoising more prone to distortions. In this paper, we set $t=50$ for all experiments.

\begin{table}[t]
  \centering
  \caption{Ablation study of dual-channel inputs on IRSTD-1K. Experiments verify the effects of background information utilization, with evaluation metrics including IoU($10^{-2}$), $P_d$ ($10^{-2}$), and $F_a$ ($10^{-6}$). ``Concat''  means the concatenation operation, and ``$\mathcal{HP}$'' denotes the high-pass filtering operation.}
  \small
  \setlength{\tabcolsep}{1.3mm}{
  \begin{tabular}{clccc}
  \toprule    
  \textsc{In-channels} & \textsc{Input design} & IoU $\uparrow$ & P$_d \uparrow$ & F$_a \downarrow$ \\
  \midrule
  2 & $\text{Concat}(x,\mathcal{HP}(x))$ & 65.47 & 89.22 & 11.06 \\
  2 & $\text{Concat}(x,x_B)$ & 72.13 & 91.91 & 13.64 \\
  \midrule
  2 & $\text{Concat}(x,|x-x_B|)$ & 70.85 & 94.61 & 15.90 \\
  2 & $\text{Concat}(x,(x-x_B)^2)$ & 71.77 & 90.90 & 13.98 \\
  3 & $\text{Concat}(x,x_B,|x-x_B|)$ & 69.53 & 93.93 & 21.12 \\
  3 & $\text{Concat}(x,x_B,(x-x_B)^2)$ & 69.45 & 93.26 & 21.92 \\
  \bottomrule    
  \end{tabular}}
  \label{tab:ablation_input_design}
\end{table}

\begin{table}[t]
  \centering
    \caption{Ablation study of frequency-guided perceptual loss coefficient $\lambda$ on IRSTD-1K. The evaluation metrics are IoU ($10^{-2}$), $P_d$ ($10^{-2}$), and $F_a$ ($10^{-6}$).}
    \small
    \setlength{\tabcolsep}{7mm}{
    \begin{tabular}{cccc }
    \toprule    
    $\lambda$ & IoU $\uparrow$ & P$_d \uparrow$ & F$_a \downarrow$ \\
    \midrule
    0.01 & 69.62 & 92.92 & 25.46  \\
    0.05 & 70.21 & 94.61 & 18.40  \\
    0.1 & 72.13 & 91.91 & 13.64  \\
    0.3 & 70.91 & 92.25 & 23.17  \\
    0.5 & 70.60 & 91.24 & 16.75  \\
    0.7 & 70.04 & 93.60 & 23.40  \\
    \bottomrule
    \end{tabular}}
  \label{tab:ablation on lambda}
\end{table}

\noindent \textbf{Designs of Dual-channel Inputs}
To effectively exploit the reconstructed background image $x_B$, we adopt a dual-channel input paradigm, where the original image $x$ is combined with auxiliary information derived from $x_B$. The key idea is that $x_B$ provides a purified background prior, which can implicitly highlight the foreground–background discrepancy and thus enhance the perception of small targets. Therefore, we conduct a series of ablation studies with different input designs, as summarized in Tab.~\ref{tab:ablation_input_design}.
 
First, among all the two-channel configurations, the simplest design 
$\text{Concat}(x,x_B)$ achieves the best overall performance, yielding the highest IoU of 72.13 and a competitive false alarm rate. This suggests that directly feeding both the original image and the reconstructed background allows the network to implicitly learn the discrepancy between foreground and background, without requiring manually engineered transformations. Such a design preserves better information and avoids introducing additional distortions.

We also compare with the commonly used high-pass filtering pre-processing, \textit{i.e.}, $\text{Concat}(x,\mathcal{HP}(x))$, which is designed to emphasize high-frequency components and suppress smooth background regions in infrared images. 
Although this strategy can highlight potential target regions, it leads to noticeably inferior performance. This indicates that handcrafted high-frequency enhancement is less effective than our reconstruction-based approach, and further demonstrates that our DiffIRSTD framework provides a more accurate and adaptive way to separate foreground from background.

In addition, we explicitly construct difference-based inputs, such as 
$|x-x_B|$ and $(x-x_B)^2$, to directly model the discrepancy. However, these variants result in slight performance degradation compared to the simple concatenation. One possible explanation is that the explicit difference map, while providing the potential target region prior, may also amplify noise or reconstruction errors, thereby misleading the detection network. This phenomenon is also observed in the studies of triple-channel inputs, and significantly increases F$_{a}$. 

\begin{table}[t]
  \centering
    \caption{Ablation study of high-pass cutoff frequency $r_c$ on IRSTD-1K. The evaluation metrics are IoU ($10^{-2}$), $P_d$ ($10^{-2}$), and $F_a$ ($10^{-6}$).}
    \small
    \setlength{\tabcolsep}{7mm}{
    \begin{tabular}{cccc }
    \toprule
    $r_c$ & IoU $\uparrow$ & P$_d \uparrow$ & F$_a \downarrow$ \\
    \midrule
    0.1 & 67.74 & 92.92 & 41.50  \\
    0.3 & 69.87 & 94.27 & 24.50  \\
    0.5 & 70.26 & 94.94 & 15.88  \\
    0.7 & 72.13 & 91.91 & 13.64   \\
    0.9 & 68.40 & 94.27 & 26.64  \\
    \bottomrule
    \end{tabular}}
  \label{tab:ablation on high-pass r}
\end{table}

\noindent \textbf{Coefficients in Frequency-guided Perceptual Loss.}
As discussed in the main paper, the frequency-guided perceptual loss is introduced to explicitly preserve the structural details of reconstructed images in the frequency domain. Results of varying loss coefficients are reported in Tab~\ref{tab:ablation on lambda}. We can observe that
\textbf{(1)} When $\lambda$ is too small, the high-frequency constraint becomes relatively weak, which may lead to insufficient preservation of structural details in the reconstructed background, slightly degrading detection performance. 
\textbf{(2)} Conversely, when $\lambda$ is too large, the high-frequency loss tends to dominate the optimization process, encouraging the network to overemphasize detail reconstruction, which may introduce noise or high-frequency artifacts and thus slightly increase false alarms.
\textbf{(3)} The best comprehensive performance is achieved at $\lambda=0.1$, where the model obtains the highest IoU (72.13) and the lowest false alarm rate. 
Overall, although the performance varies with different $\lambda$, the fluctuations are not significant, and the model consistently achieves competitive results. This suggests that the proposed frequency-guided loss is inherently stable, while a moderate setting provides the best trade-off between structural consistency and detail fidelity, leading to optimal detection performance.

\noindent \textbf{Choice of High-pass Cutoff Frequency $r_c$.}
It can be observed from the experimental results in Tab~\ref{tab:ablation on high-pass r} that the cutoff radius coefficient $r_c$ of the high-pass filter exerts a crucial impact on the detection performance. The parameter $r_c$ controls the retention range of high-frequency components in the frequency domain. If $r_c$ is too small, the high-frequency constraint becomes excessively lenient, causing the generated background to lose a large number of edge details, which in turn leads to a low IoU and a high false alarm rate. If $r_c$ is too large, low-frequency information is overly suppressed, and the generated background may produce texture artifacts or structural distortion, also resulting in performance degradation. The experimental results demonstrate that $r_c = 0.7$ can achieve the optimal balance between retaining necessary high-frequency details and maintaining the overall structure of the background, thereby maximizing the detection performance.

\begin{table*}[t]
  \centering
    \caption{Comparisons with different reconstruction methods. The evaluation metrics are IoU ($10^{-2}$), $P_d$ ($10^{-2}$), and $F_a$ ($10^{-6}$).}
    \setlength{\tabcolsep}{4mm}{
    \begin{tabular}{cccccccccc }
    \toprule
    \multicolumn{1}{c}{\multirow{2}{*}{\textsc{Reconstruction Network}}} & \multicolumn{3}{c}{\textbf{IRSTD-1K}}& \multicolumn{3}{c}{\textbf{NUDT-SIRST}}& \multicolumn{3}{c}{\textbf{NUAA-SIRST}} \\
    \cmidrule(r){2-10}
    & IoU $\uparrow$ & P$_{d} \uparrow$ & F$_{a} \downarrow$ & IoU $\uparrow$ & P$_{d} \uparrow$ & F$_{a} \downarrow$ & IoU $\uparrow$ & P$_{d} \uparrow$ & F$_{a} \downarrow$\\
    \midrule
    Auto-Encoder & 67.01 & 90.23 & 15.97 & 94.95 & 99.36 & 2.57 & 78.99 & 94.29 & 10.44 \\
    One-Step Diffusion & 72.13 & 91.91 & 13.64 & 95.91 & 99.49 & 1.72 & 80.73 & 96.95 & 7.55   \\
    \bottomrule
    \end{tabular}}
  \label{tab:autoencoder}
\end{table*}

\begin{table}[t]
  \centering
  \caption{Generalization of our DiffIRSTD with other detectors. The red number means the performance gain on IRSTD-1K.}
  \small
  \setlength{\tabcolsep}{2.5mm}{
  \begin{tabular}{lccc}
  \toprule
  \multicolumn{1}{c}{\textsc{Methods}} & IoU $\uparrow$ & P$_d \uparrow$ & F$_a \downarrow$ \\
  \midrule
  ACM~\cite{dai2021asymmetric} & 60.98\textcolor{white}{$_\text{+0.00}$} & 89.89\textcolor{white}{$_\text{+0.00}$} & 38.26\textcolor{white}{$_\text{+0.00}$} \\
  \rowcolor{gray!10} \quad + DiffIRSTD (\textit{Ours}) & 64.71\textcolor{red}{$_\text{+3.73}$} & 90.90\textcolor{red}{$_\text{+1.01}$} & 36.00\textcolor{red}{$_\text{-2.26}$} \\
  DNANet~\cite{li2022dense} & 65.50\textcolor{white}{$_\text{+0.00}$} & 90.23\textcolor{white}{$_\text{+0.00}$} & 18.80\textcolor{white}{$_\text{+0.00}$} \\
  \rowcolor{gray!10} \quad + DiffIRSTD (\textit{Ours}) & 69.53\textcolor{red}{$_\text{+4.03}$} & 91.24\textcolor{red}{$_\text{+1.01}$} & 14.59\textcolor{red}{$_\text{-4.21}$} \\
  SCTransNet~\cite{yuan2024sctransnet} & 67.28\textcolor{white}{$_\text{+0.00}$} & 95.28\textcolor{white}{$_\text{+0.00}$} & 19.49\textcolor{white}{$_\text{+0.00}$} \\
  \rowcolor{gray!10} \quad + DiffIRSTD (\textit{Ours}) & 70.04\textcolor{red}{$_\text{+2.76}$} & 95.58\textcolor{red}{$_\text{+0.30}$} & 18.95\textcolor{red}{$_\text{-0.54}$} \\
  \bottomrule
  \end{tabular}}
  \label{tab:different detector}
\end{table}

\noindent \textbf{Generalization to Other Detectors.}
Table~\ref{tab:different detector} evaluates the generalization ability of our DiffIRSTD framework when combined with different detection networks. We reproduce several representative works \cite{dai2021asymmetric,li2022dense,yuan2024sctransnet} with the official\footnote{SCTransNet: \url{https://github.com/xdFai/SCTransNet}} or popular\footnote{BasicIRSTD: \url{https://github.com/XinyiYing/BasicIRSTD}} codebases.
It can be observed that our method can be seamlessly integrated with various detectors, including ACM, DNANet, and SCTransNet, demonstrating good compatibility and flexibility. With the introduction of the reconstruction module, all detectors show consistent performance improvements. Specifically, ACM achieves an IoU gain of $+3.73$, DNANet improves by $+4.03$, and SCTransNet gains $+2.76$. Similar trends can also be observed in the false-alarm rate $F_a$, where detectors benefit from noticeable reductions. This suggests the plug-and-play advantage of our DiffIRSTD framework, which provides effective background priors that can be readily exploited by different detectors, leading to consistent and stable performance improvements.

\noindent \textbf{Discussion on Reconstruction Methods.}
To validate the effectiveness of the reconstruction network, we compare two representative denoising paradigms, \textit{i.e.}, the classical auto-encoder and our one-step diffusion model, as shown in Tab.~\ref{tab:autoencoder}. Auto-encoders are widely adopted for image reconstruction and denoising due to their simple architecture and efficient inference, making them a strong baseline for background modeling. In contrast, diffusion models demonstrate superior generative capabilities, particularly in modeling complex data distributions and producing high-fidelity reconstructions. In this work, we further adopt a one-step denoising strategy to retain the strong generative prior of diffusion models while ensuring computational efficiency.

From the quantitative results, the one-step diffusion model consistently outperforms the auto-encoder across all three datasets. Specifically, it achieves higher IoU and $P_d$, while reducing the false alarm rate $F_a$. For example, on IRSTD-1K, the IoU improves from 67.01 to 72.13, accompanied by a noticeable reduction in false alarms. Similar performance gains can also be observed on NUDT-SIRST and NUAA-SIRST, demonstrating the robustness of the improvement.
These results indicate that the diffusion-based reconstruction provides a cleaner background reconstruction, which is crucial for highlighting foreground–background discrepancies in infrared small target detection. Compared with the auto-encoder, the stronger generative capacity of the diffusion model enables better denoising of target-related signals during reconstruction, thereby facilitating subsequent detection.

\section{Broader Impact and Limitation}
\subsection{Broader Impact}
In the field of infrared small target detection, this work proposes an explicit decomposition paradigm of ``background reconstruction + target detection''. In critical scenarios such as maritime search and rescue and missile early warning, this work could improve the reliability of the system with a lower false alarm rate and higher detection accuracy. Furthermore, this work promotes the application of generative models in the remote sensing field, utilizing diffusion models to achieve ``target erasure''. This idea can be migrated to other remote sensing tasks (\textit{e.g.}, cloud removal, occluder elimination), expanding a new perspective for the role of generative models in visual tasks.

\subsection{Limitations and Future Work}
Although the framework proposed in this paper achieves significant performance improvements in the infrared small target detection task, it still has some limitations. In this work, diffusion sampling is compressed to a single step, which greatly improves the inference speed and reduces the inference overhead, but in terms of computational cost, the reconstruction network still contains a large number of parameters. For practical applications, a viable solution is to deploy the reconstruction network independently in the cloud, serving as a preprocessing step during the inference process. However, if the entire framework needs to be deployed on edge devices, further lightweight optimization is still required.
An interesting future work is to investigate more computationally efficient reconstruction mechanisms and explore the applicability of this framework to broader small-object detection scenarios.

\section{Conclusion}
This paper presents DiffIRSTD, a diffusion-driven framework for infrared small target detection that approaches the problem from a reconstruction-oriented perspective. Rather than directly inferring target regions, we first recover a target-free background representation by suppressing target-related signals through a diffusion-based denoising process. The comparison differences between the reconstructed background and the original image thus naturally reveal candidate target regions, providing informative guidance for detection.
Experiments on multiple benchmarks verify the effectiveness and robustness of the proposed framework.
Furthermore, the modular design of the reconstruction component allows it to be readily integrated with different detection architectures, demonstrating strong adaptability and practical value.

\bibliographystyle{IEEEtran} 

\bibliography{main} 

\vfill

\end{document}